\documentclass[journal]{IEEEtran}

\usepackage{graphicx}
\usepackage{amsfonts}
\usepackage{amsmath}
\usepackage[small,tight]{subfigure}
\usepackage{balance}
\usepackage{multirow}
\usepackage{booktabs}

\begin{document}
	
\title{Fine-Grained Land Use Classification at the City Scale Using Ground-Level Images}
%
%
%

\author{Yi~Zhu, Xueqing~Deng, and Shawn~Newsam
\thanks{This work was supported in part by a National Science Foundation (NSF) CAREER grant, No. IIS-1150115, and a seed grant from the Center for Information Technology in the Interest of Society (CITRIS). Any opinions, findings, and conclusions or recommendations expressed in this material are those of the authors and do not necessarily reflect the views of NSF and CITRIS. We gratefully acknowledge the support of NVIDIA Corporation through the donation of the Titan X GPUs used in this work.}
\thanks{Yi Zhu, Xueqing Deng, and Shawn Newsam are with the Department
of Electrical Engineering and Computer Science, University of California, Merced, Merced, CA, 95343 USA. (email: \{yzhu25, xdeng7, snewsam\}@ucmerced.edu)}}
\maketitle

\begin{abstract}
We perform fine-grained land use mapping at the city scale using ground-level images. Mapping land use is considerably more difficult than mapping land cover and is generally not possible using overhead imagery as it requires close-up views and seeing inside buildings. We postulate that the growing collections of georeferenced, ground-level images suggest an alternate approach to this geographic knowledge discovery problem. We develop a general framework that uses Flickr images to map $45$ different land-use classes for the City of San Francisco. Individual images are classified using a novel convolutional neural network containing two streams, one for recognizing objects and another for recognizing scenes. This network is trained in an end-to-end manner directly on the labeled training images. We propose several strategies to overcome the noisiness of our user-generated data including search-based training set augmentation and online adaptive training. We derive a ground truth map of San Francisco in order to evaluate our method. We demonstrate the effectiveness of our approach through geo-visualization and quantitative analysis. Our framework achieves over $29\%$ recall at the individual land parcel level which represents a strong baseline for the challenging $45$-way land use classification problem especially given the noisiness of the image data.
\end{abstract}

\begin{figure}[t]
	\centering
	\includegraphics[width=0.90\linewidth,trim=0 0 0 0,clip]{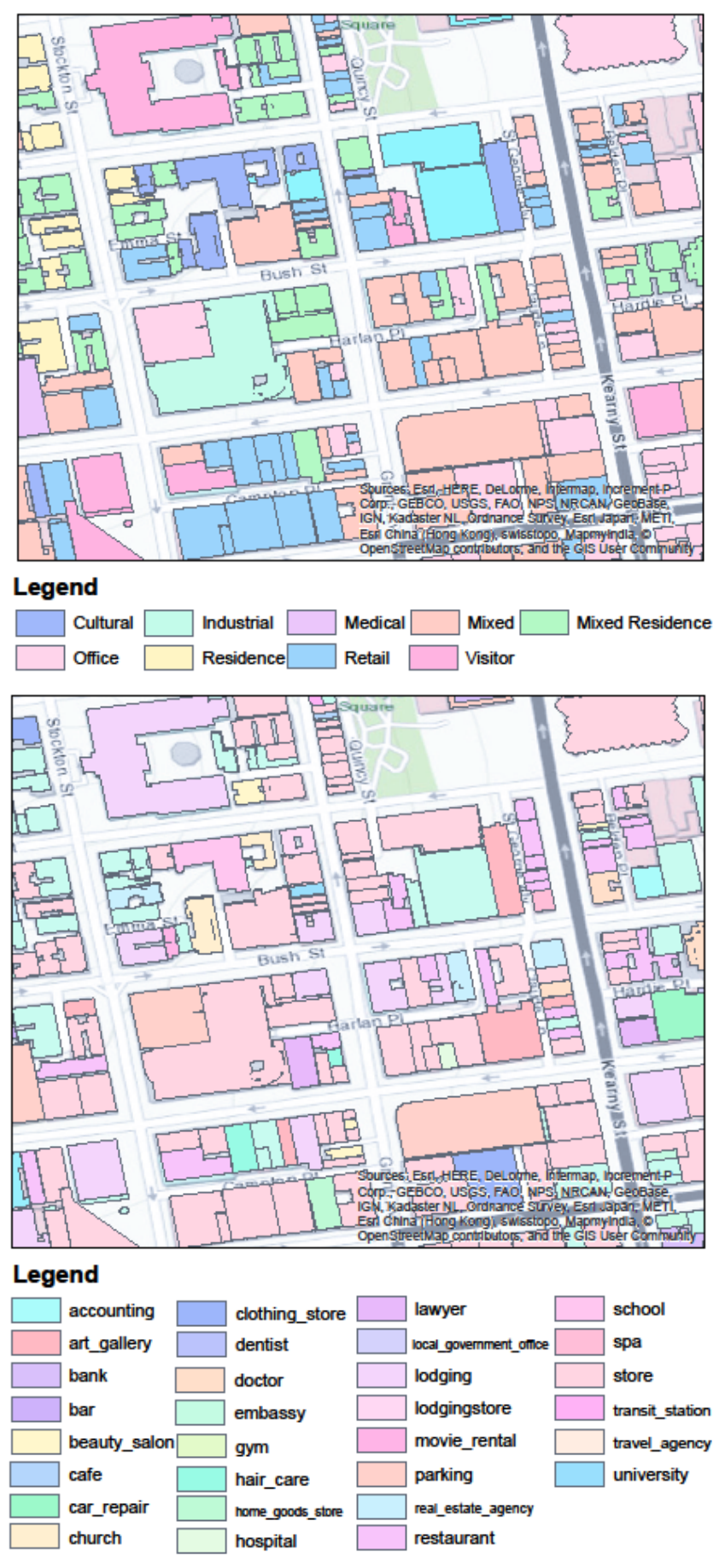}
	\vspace{-2ex}
	\caption{Fine-grained land use map. The two maps are for the same location with different annotations. Upper: Traditional map with less than $10$ land use types. Bottom: our ground-truth map with fine-grained classes. The figure is best viewed in color. }
	\label{fig:fine_grain}
	\vspace{-2ex}
\end{figure}


%
\IEEEpeerreviewmaketitle

\section{Introduction}
\label{sec:introduction}

\IEEEPARstart{D}{etailed} and accurate land use information is of substantial importance towards building a smart city \cite{smart_city_Yamagata_J13,urban_plan_Llacuna_J14,monitor_Rawat_J15,PSOLA_Liu_PLOS16}, such as environmental monitoring, urban planning, resource allocation, traffic control and governmental management. Especially the transformation of land use patterns over time contains a wealth of information for both the government and individuals to make informative decisions. However, traditional land use map is generated using survey-based approaches, which cost enormous human effort. Such map is usually updated for every $5$ to $10$ years, thus losing the important information of how the urban functional structures are changing. Hence, it is significant to develop a system which can automatically generate accurate and up-to-date land use map on large-scale. 

Most literature on land use classification resort to high-resolution remote sensing images \cite{landuse_cnn_Castelluccio2015,transfer_Zhao_GRS17,bayesian_Li_IEEEJ17}. It might be easy to distinguish airport from residential area \cite{bow_yiyang_sigspatial10} just using overhead imagery. However, it is much more difficult to determine land use in complicate urban areas from above. Images taken at ground-level are potentially more indicative. For example, to determine a building whether it is a restaurant or a barber shop, overhead imagery can't help at all whereas a glance of ground-level images will give you the right answer. Thus, using aerial images has a limitation that it is infeasible to generate a fine-grained land use map, which is referred as the ``semantic gap'' problem. 
Some recent efforts \cite{landsat_osm_Hu_RS17,deng_sigspatial17} propose to use other data sources that can bring more information about the internal structure or human activities of the land to close the semantic gap, such as point of interest (POI) \cite{poi_JIANG_J2015}, street view \cite{street_view_Kang_ISPRS17}, mobile phone data \cite{mobile_Pei_IJGIS14} and social multimedia \cite{remote_social_Liu_IJGIS17}. However, approaches that depend on POI, street view, mobile records and textual tweets may also face the problem that we have limited observation inside the building. 

Hence, in this work, we explore the rich online photo collections to perform large-scale fine-grained land use mapping. Popular social photo sharing websites such as Facebook, Twitter, Instagram, Pinterest, Flickr and Picasa present an all-around view of the world and contain a wealth of information. With more than $400$ million geotagged images on Flickr alone, there is an opportunity to automatically generate up-to-date city-scale land use map. However, we still face many challenges as below:
\begin{itemize}
	\item Due to the lack of both ground truth map and training data, there is no standard benchmark to evaluate land use classification methods.
	\item Online photo collections are too large to be manually labelled, which means that weakly supervised or unsupervised learning methods are preferred.
	\item Ground-level web images are too noisy to be mined. Challenges include low image quality, inaccurate geotags and uneven spatial distribution, etc. 
\end{itemize}

To address these challenges, we first introduce a ground truth land use map of the city of San Francisco, which could be used as a benchmark to evaluate various approaches. The ground truth map has a three level hierarchy: $5$ top classes, $16$ middle classes and $45$ fine-grained classes. We then train deep Convolutional Neural Network (CNN) models to learn to classify land use in an end-to-end manner. We propose several strategies to overcome the noisiness of online photos, like search-based dataset augmentation, online adaptive training, and two-stream networks of object and scene.

\begin{figure}[t]
	\centering
	\includegraphics[width=1.0\linewidth,trim=0 50 0 0,clip]{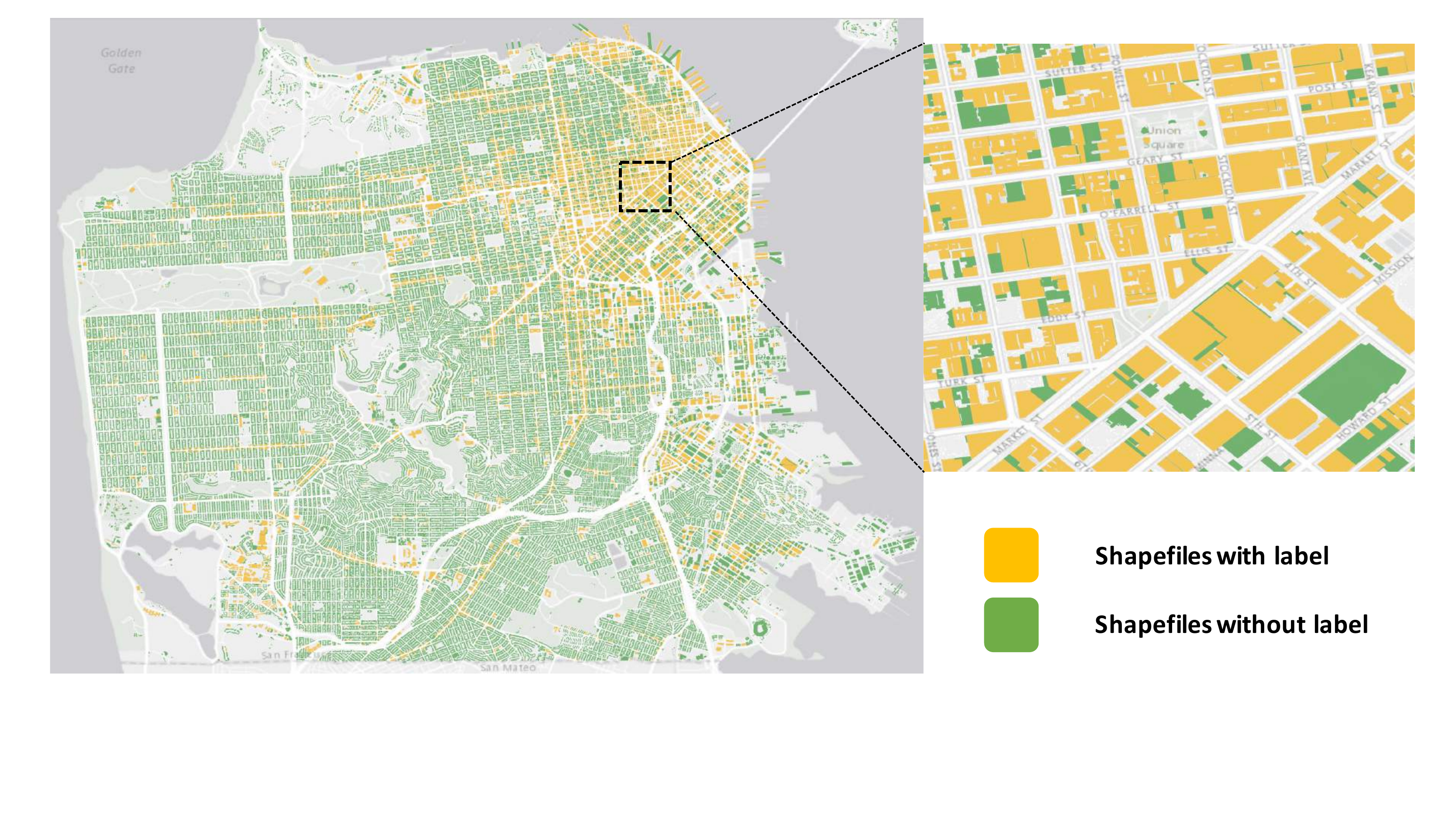}
	\vspace{-4ex}
	\caption{We introduce a fine-grained land use map of the city of San Francisco with ground truth. Right figure is the zoomin view of the black dashed box in the left figure. Yellow indicates a shapefile with ground truth label, and green without. Note the sparsity (about $10\%$ shapefiles have ground truth). The figure is best viewed in color. }
	\label{fig:problem_overview}
\end{figure}

Our work in this paper represents a thorough investigation into mapping fine-grained land use types on large-scale along with novel algorithmic contributions. The most distinctive point between previous literature and our work is the \textit{fine granularity}. As we show in Figure \ref{fig:fine_grain} that traditional work usually perform coarse-level land use classification, e.g., the number of land use types are less than $10$. However, we aim to address the fine-grained land use classification problem, e.g., our dataset has $45$ land use types and is easily extended to more classes. The fine granularity requirement makes the problem more challenging yet more appealing. The salient contributions of our work include:
\begin{itemize}
	\item To the best of our knowledge, our work is the first to conduct fine-grained land use mapping. The number of land use types in finest granularity, $45$, is much more diverse than previous literature. 
	\item We combine the Land Based Classification Standards (LBCS) and Google places API to create a ground truth map of the city of San Francisco. Although sparse (see in Figure \ref{fig:problem_overview}), it can serve as a benchmark to evaluate various approaches using different data sources for land use classification.
	\item We introduce online adaptive training technique to reduce the impact of noisy web images during end-to-end model learning. The strategy not only increases the accuracy of land use classification, but also makes our trained model more robust for domain adaptation. 
	\item We propose to use two-stream networks, consists of an object-centric model and a scene-centric model to further improve the performance. 
\end{itemize}

Our paper is organized as follows. After introducing related work in Section \ref{sec:relatedwork}, we illustrate how we build the ground truth land use map and the details of our land use classification framework in Section \ref{sec:fine_grained_land_use_classification}. In Section \ref{sec:experiments}, we describe our dataset, implementation details, results and geo-visualizations. We then investigate the design options and discuss the mapping results of our proposed method in Section \ref{sec:discussion}, and present our conclusion in Section \ref{sec:conclusion}. 

\section{Related Work}
\label{sec:relatedwork}
\noindent Our work has several lines of related research.

\noindent \textbf{Large-scale geotagged photo collections} Computer vision researchers have been leveraging large collections of geotagged photos for geographic discovery for around a decade. This includes mapping world phenomenon \cite{CrandallMapWorld09}, multimedia geolocalization \cite{Haysim2gps08}, landmark recognition and 3D modeling \cite{3dModelSnavely08}, smart city and urban planning \cite{smartUrban15}, land cover and land use classification \cite{landuse_cnn_Castelluccio2015}, sentiment hotspot detection \cite{emotion_yi_sigspatial_2016} and mapping human activity \cite{activity_yi_sigspatial_2017}. The exponential growth of photo sharing and related websites along with ever more publicly available multimedia sources make this research paradigm a promising direction for a range of interesting problems. Although such open-access multimedia represents a wealth of information, analyzing it is challenging due to how noisy and diverse it is. Challenges specific to using this data for geographic discovery include inaccurate location information, uneven spatial distribution, varying photographer intent and license limitations. We are mindful of these challenges and recognize they likely temper our results.

Our work is novel in that it uses a large collection of geotagged photos to perform fine-grained land use classification. We address several challenges mentioned above. We build a ground truth map using Google places data for proper evaluation. We also use region shapefiles to reduce geolocation error. In addition, we create a very large training set (more than two million images) consisting of Google and Flickr images to learn a robust CNN model.

\noindent\textbf{Convolutional neural networks} Deep learning is advancing a number of pattern recognition and machine learning areas. Deep convolutional neural networks (CNNs) have resulted in often surprising performance gains in a range of computer vision problems \cite{AlexImageNet12,girshick14CVPR,KarpathyCVPR14}. Key to CNNs performance is their ability to learn high-level or semantic features from the data as opposed to the hand-crafted low- to mid-level features traditionally used in image analysis. Visualization of the feature maps learned by the convolutional layers during training \cite{Zeiler_visualizeCNN_ECCV14} shows how the features become increasingly semantic, progressing from pixels, edges, color, and texture, to motifs, parts, objects, scenes, and concepts. Another significant benefit of the features learned by CNNs is their ability to generalize to problems involving image datasets other than the ones they were trained on \cite{midlevel14}. This avoids having to retrain the networks which can take from hours to days even on powerful GPUs. Hence, many work adopt deep learning to advance the state-of-the-art in land use classification \cite{patch_Zhong_J16,ft_generalize_Penatti_CVPR15,extreme_Weng_GRS17,Multiview_Luus_GRS15,Repurposing_Tracewski_J17,imagenet_Marmanis_GRS16,EuroSAT_Helber_17}.

In this paper, we also use CNN to classify land use but introduce a novel learning technique named online adaptive training to reduce the impact of noisy web images during model fine-tuning. Besides, we propose to use two-stream networks, consists of an object-centric model and a scene-centric model to further improve the land use classification performance.

\noindent\textbf{Land cover and land use classification} Land cover and land use classification are important tasks in geographic science. The maps they produce are critical for a range of important societal problems. However, land cover is distinct from land use, despite the two terms often being used interchangeably. Land cover is the physical material at the surface of the earth, which includes grass, trees, bare ground, water, etc. Land use is a description of how people utilize the land and of social-economic activity. Land cover classification is typically performed through the automated analysis of overhead imagery. However, land use classification is more difficult since it is often not possible from an overhead vantage point. We need to see inside buildings to determine their use(s). We also need to resolve details which are not discernible in today's overhead imagery or are only apparent from ground-level.

Researchers have performed some initial investigation into using ground-level photo collections for land cover \cite{landcoverFlickr14,landuse_US_Theobald_PLOS14} and land use \cite{Shekhar_TMM02,Danielproximate10,landuse_yi_sigspatial_2015} classification. Here, we only consider land use classification problem. \cite{Danielproximate10} has only considered a two-class land use problem: developed and undeveloped, and \cite{landuse_yi_sigspatial_2015} has considered a limited number of land use classes in an university campus. Both of them avoid the challenge of lacking ground truth land use map since they can use the off-the-shelf city zoning map and campus map. However, when we turn to fine-grained land use classification on large-scale instead of these toy examples, we could not avoid these challenges. There also exists some work \cite{finegrain_Untenecker_LUpolicy16,nazrcnn_Attari_J16,finegrain_Zhang_RS17} focusing on the concept of ``fine-grained'' in land use classification. However, \cite{finegrain_Untenecker_LUpolicy16} indicates fine granularity on time scale and \cite{nazrcnn_Attari_J16} indicates fine granularity on levels of damage. None of them refer to the granuality of land use types.

Hence, in this paper, we create a ground truth land use map of the city of San Francisco to serve as a benchmark for evaluating various approaches. In addition, the number of our land use types, $45$, is a magnitude more than previous work. 

\begin{figure*}[t]
	\centering
	\includegraphics[width=1.0\linewidth,trim=0 80 0 10,clip]{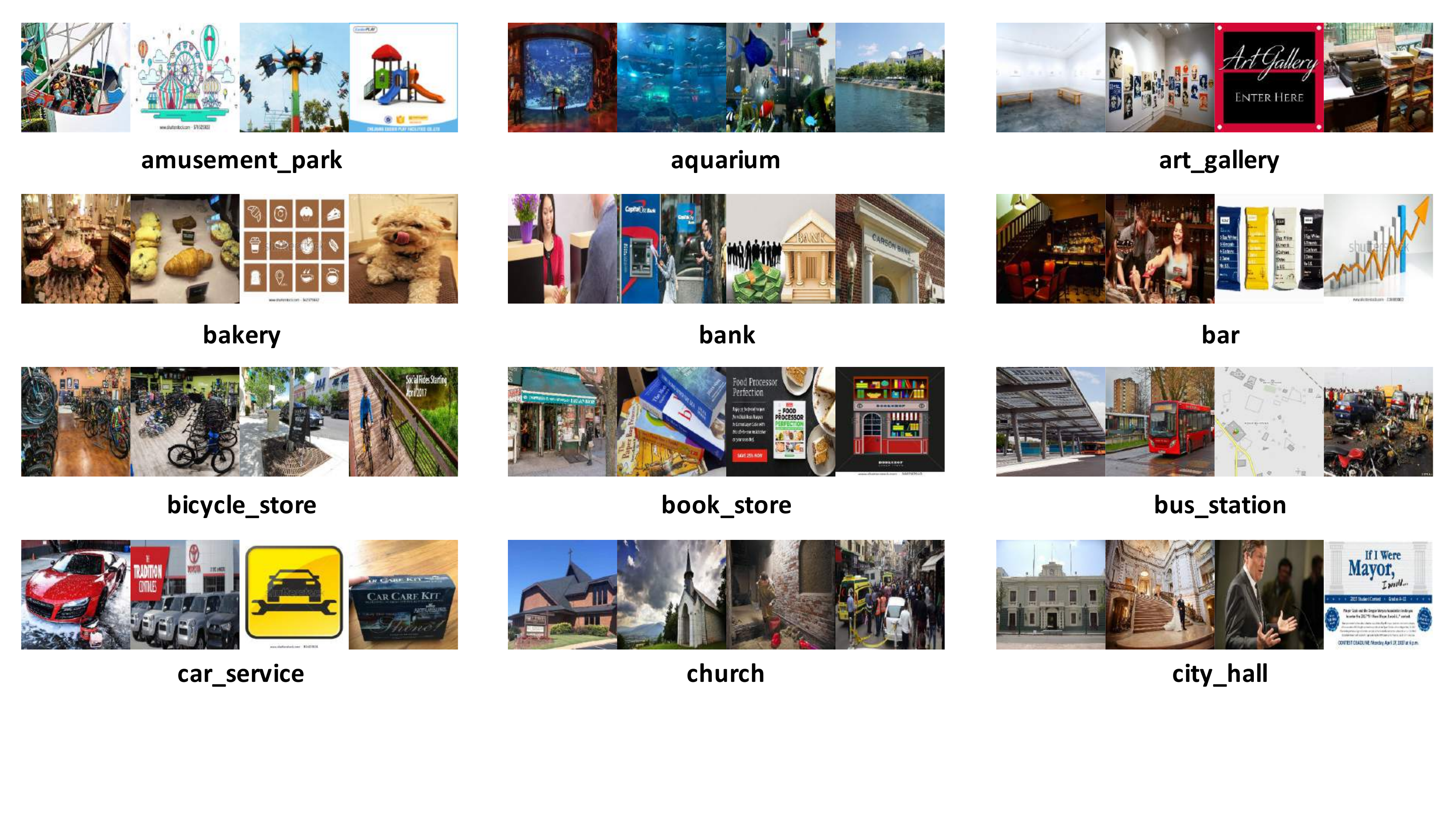}
	\caption{Sample images from our training dataset. For each class, we display $4$ images. The images are arranged in the order of closeness to that land use type from left to right. Note the noisiness contained in the web-crawled dataset. The figure is best viewed in color. }
	\label{fig:sample_images}
\end{figure*}

\section{Fine-Grained Land Use Classification}
\label{sec:fine_grained_land_use_classification}

In this section, we focus on fine-grained land use classification. The details of the dataset construction is first illustrated in Section \ref{subsec:construction_of_the_database}. In Section \ref{subsec:end_to_end_learning}, we describe our end-to-end learning framework. Then we introduce the online adaptive training strategy to reduce the impact of noisy web images during model fine-tuning in Section \ref{subsec:online_adaptive_training}. Finally, we propose the two-stream networks, consisting of an object-centric and a scene-centric model to further improve the land use classification accuracy in Section \ref{subsec:two_stream}. 

\begin{figure*}[t]
	\centering
	\includegraphics[width=1.0\linewidth,trim=20 0 20 0,clip]{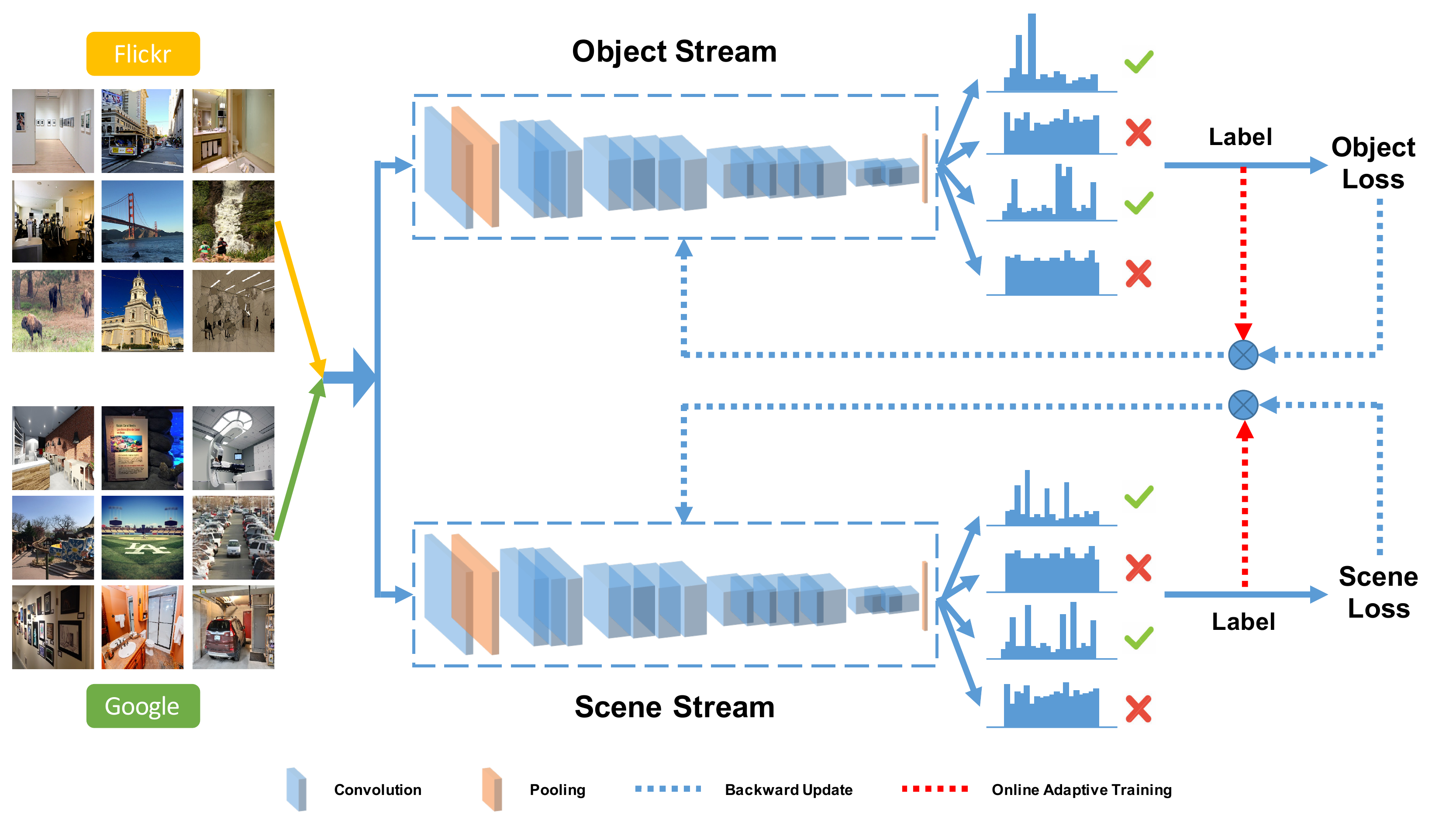}
	\caption{Overview of our fine-grained land use classification framework. The input are batches of mixed images downloaded from Flickr and Google. We train two CNN models to learn complementary object information and scene information. During inference, their prediction scores are fused with equal weights. The online adaptive training strategy is outlined in red dashed lines during backward model weights update, to reduce the impact of hard examples. The figure is best viewed in color. }
	\label{fig:framework_overview}
\end{figure*}

\subsection{Construction of the Database}
\label{subsec:construction_of_the_database}

\noindent \textbf{Database Structure}

\noindent In this work, we aim to deal with real-world land use classification problem (a.k.a, zoning problem). The first challenge we are facing is that we need to design a ground truth database for both training and evaluation. The database should have a well-defined structure with multiple hierarchies, consisting of diversified and meaningful land use categories.

Previous literature adopt two ways to avoid building the ground truth database. One is manually define the land use classes in a controlled environment such as university campus \cite{landuse_yi_sigspatial_2015}, where the land use map can be easily derived. The other is to use coarse zoning information provided by government to train and evaluate their models.

Here, we construct the database in a principled way by following the LBCS\footnote{More detailed information of the LBCS standard can be found in https://www.planning.org/lbcs/}: a consistent model for classifying land uses based on their characteristics. The LBCS model extends the notion of classifying land uses by refining traditional categories into multiple dimensions, such as activities, functions, building types, site development character, and ownership constraints. Each dimension has its own set of categories and subcategories. These multiple dimensions allow users to have precise control over land use classifications. 
Here, we pick the Function dimension. ``Function'' refers to the economic function or type of establishment using the land, which is the natural way of doing land use classification.

However, directly using the LBCS Function standard is infeasible because we don't have a corresponding ground truth to evaluate the performance of our method. As mentioned in related work, it takes enormous manual effort to come up with the ground truth for such large-scale land use classification problem. Fortunately, Google places API\footnote{https://developers.google.com/places/} provides us a list of places types, most of those correlate well with land use. 
Although the Google records are spatially sparse (see in Figure \ref{fig:problem_overview}), they are relatively accurate and updated every year, which can be regarded as ground truth for our problem at hand. 
Thus we could make a well-defined ontology with ground truth by combining LBCS Function standard and Google places API. 

Since we aim to classify land use using only user contributed social multimedia (e.g., Flickr images), we face the challenge that for some land use types, people rarely take photos. For example, ``atm'' due to privacy issues. And some places types from Google places are not indicating land use, such as ``roofing$\_$contractor''. Hence, we remove the classes that are (1) without enough online photos; (2) not land use indicative and (3) have no ground truth from Google places.

In this manner, we obtain the final database structure. It is a 3-level hierarchy, with $5$ top classes, $16$ middle classes and $45$ bottom classes. The five top classes are: (1) Residence or accommodation functions; (2) General sales or services; (3) Transportation, communication, information, and utilities; (4) Arts, entertainment and recreation; (5) Education, public admin, health care and other institution. The full hierarchy is given in the Appendix A.

There are at least two benefits of having such annotated land use map\footnote{The annotated land use map will be released in GIS compatible format to facilitate research in this field.}. First, researchers could use our map to evaluate their methods for land use/cover classification, no matter what the inputs are. For example, they can use Twitter texts, Instagram images, YouTube videos, remote sensing images, and even sensor signals from internet-of-things. Second, a good benchmark can enable fair comparison and facilitate research in this field. 

\vspace{2ex}
\noindent \textbf{Training and Mapping Set}

\noindent Once the ontology is ready, we start to collect our dataset for training deep CNN models to classify land use. 

The mapping set, which is actually the test set, consists of the geotagged Flickr images from San Francisco region of year 2016. We download a total of $96,382$ images. Due to the noisy geotags and uneven spatial distribution, we would like to keep as many photos as possible to do the final mapping for better geo-visualization. But how to come up with a large and diverse training set?  
We note, however, that we do not need location information for the training images, all we need is label information. We thus propose a search based approach to build the training set. 

Specifically, we perform keyword search on Google images. The keyword is the name of each bottom class. To increase the size of training set, we perform multiple related keyword searches for each category. For example, for bottom class ``school'', we could use other keywords such as ``elementary school'', ``high school'', ``adult school'', etc. We find the top retrievals are largely relevant. This strategy results in a total of $35,478$ images for a single category ``school''. And following this method, we build an initial training set with more than $1$ million images. 
 
However, there exists one concern that since we will use Flickr images to do the final mapping, the domain gap between Google images and Flickr images may harm the generalization of our trained model. Most Google images have clean background, whereas most Flickr images have large faces or humans, have been photoshoped or have complex background.  
Hence, we perform another keyword search on Flickr to augment the initial training set. Note that, in order to avoid duplicates in training and test set. We download these Flickr images from other cities far away from San Francisco, like Paris, Atlanta, New York, Dallas etc. 

In the end, our final training set consists of over $2$ million images. We randomly split the dataset with a ratio of $0.8$ to $0.2$ for training and validation of our CNN model. Sample images of our crowd-sourced dataset can be seen in Figure \ref{fig:sample_images}. 

This search based strategy has three benefits: (1) it results in a balanced, rich training set; (2) it preserves all the images in San Francisco with location information for mapping; and (3) the data labeling procedure is efficient and automated. Although we may concern that using web data may introduce additional label noise, many literature demonstrate that certain level of noise doesn't impact the performance of deep neural network, and sometimes even improve the generalization ability of the trained model \cite{DisturbLabel_xie_cvpr16}. 

\vspace{2ex}
\noindent \textbf{Geo-filtering with shapefiles}

\noindent We use the polygonal outlines of the land use regions we want to classify to filter noisy images and to produce more precise maps. These irregularly shaped polygons are known as shapefiles in geographical information systems (GIS) and are widely available. Figure \ref{fig:problem_overview} shows the shapefiles for our problem. Using the shapefiles has two benefits:

\begin{itemize}
	\item Filtering: We ignore the images which do not fall in one of the regions we want to classify. In our experiment, in order to tolerate the geolocation error of web images, we dilate the shapefiles. We regard a photo that is $5$ meters away from the region boundary is still within this region. This removes a lot of noisy (unrelated) images and reduces our dataset from $96,382$ images to $58,418$.
	\item Precision: The ground truth land use map in Figure \ref{fig:problem_overview} was generated using the shapefiles. It is very precise and could be published with very few modifications such as overlaying the street network.
\end{itemize}

Compared with the tiling and discretization approach of previous work \cite{DanielLandUse12,geographUK,NCLD2006accuracy13}, incorporating shapefiles results in maps which are more map-like. They are significantly more geo-informative visually.

\begin{figure*}[t]
	\centering
	\includegraphics[width=1.0\linewidth,trim=0 0 0 0,clip]{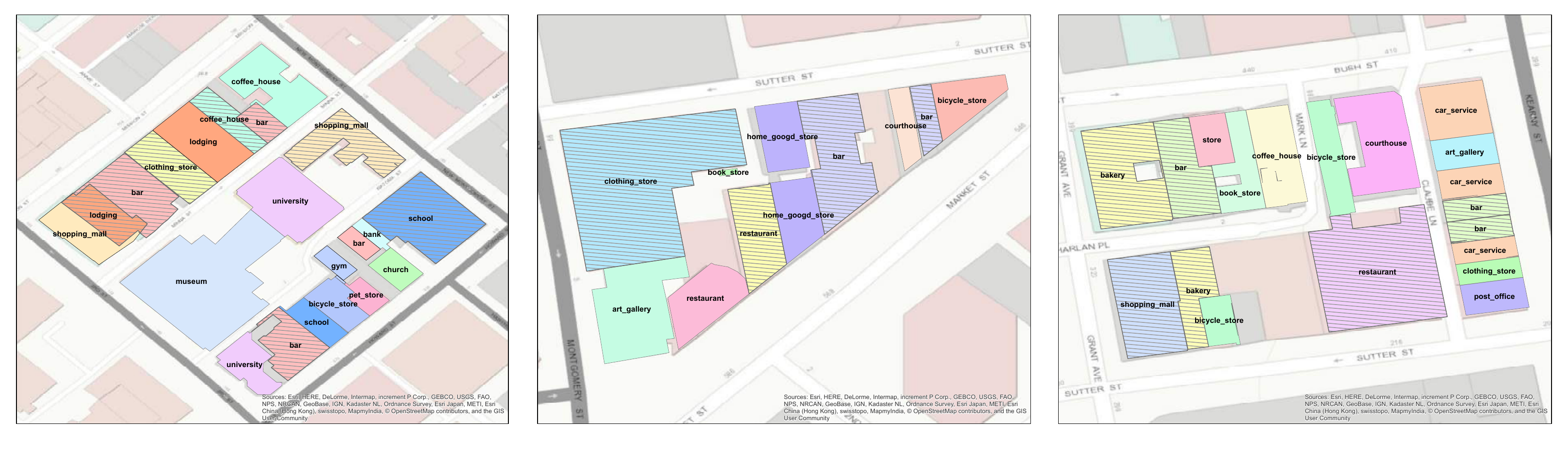}
	\vspace{-2ex}
	\caption{Sample geo-visualizations in downtown San Francisco. Though each shapefile (building) may contain multiple land use types, we show the one with majority votes for clear visualization. Slashed regions are correct predictions with respect to Google Map. The figure is best viewed in color.}
	\label{fig:geo_visualizations}
\end{figure*}

\subsection{End-to-End Learning}
\label{subsec:end_to_end_learning}

Most traditional approaches for land use/cover classification adopt a two stage pipeline \cite{bow_yiyang_sigspatial10}, which first extract image features (like color histogram, shape, texture, Scale Invariant Feature Transform (SIFT), GIST, or deep CNN features) and then train an image classifier (like logistic regression, support vector machine (SVM), or shallow neural network). This two stage framework has the following shortcomings: (1) The two stages are independent of each other, hence the solution may stuck in local minimal, and can not achieve a global optimal solution. (2) Image features need to be cached, at least during training stage, which is computational and storage forbidden for large-scale applications.

Recent approaches use an end-to-end learning strategy with deep neural networks. However, most land use/cover literature focus on remote sensing images, rare work use social multimedia. The challenges mainly come from two perspectives, one is the lack of ground truth dataset to evaluate, the other is the noisiness and uneven spatial distribution of the online photo repositories. 

To overcome these challenges, in this work, we collected a weakly annotated dataset with more than $2$ million images, distributed over $45$ land use types, and organized in a three-level hierarchy. Such large-scale dataset could compensate the noisiness of online images and be used to train an effective deep CNN model for land use classification. 

Training a CNN model is straightforward. We use the keywords as image labels, and train a $45$-way classifier. One thing to notice is that, in order to cross the domain gap between Google images and Flickr images, our input to the CNN is comprised of images from both domains. To be specific, we use a batch size of $256$ during training, half of which is from Google images, and the other half comes from Flickr. We adopt ResNet101 \cite{resnet_CVPR16} as our network architecture due to its good trade-off between accuracy and efficiency. We also explore other model architectures, like AlexNet \cite{AlexImageNet12}, VGG16 \cite{vgg16192015}, GoogleNet \cite{googleNet2015}, ResNet34 \cite{resnet_CVPR16} and DenseNet121 \cite{huang2017densely}. The implementation details can be seen in Section \ref{sec:implementation}. We report their performance on our validation set with discussion in Section \ref{sec:cnn_architecture_search}.

\subsection{Online Adaptive Training}
\label{subsec:online_adaptive_training}
Using deep CNN models, especially their pretrained models \cite{imagenet_CVPR09}, we can leverage the powerful learned representations to adapt to many tasks and achieve promising results. However, such transfer learning require the problem is well-defined and the fine-tuning dataset at hand is also large and clean. 

As for our problem, the social multimedia is labelled by keywords search, not manually annotated. Such weakly labelled dataset will lead to bad local minimum or model collapse during training. Manually cleaning the dataset would be ideal, but it is unrealistic with respect to the exponential growth of social multimedia. Thus, we need to tolerate the noisiness and try to learn useful visual representation from weakly annotated samples, similar as the recent released large-scale dataset, WebVision \cite{webvision} and YouTube-8M \cite{YouTube_8M_2016}. 

Noted that, even the dataset is noisy, our directly trained CNN achieves moderate accuracy as seen in Table \ref{tab:results_tab1}. Therefore, we propose to use the trained model to perform an online adaptive training, as an unsupervised dataset cleaning procedure. This strategy is used to reduce the impact of noisy instances during model fine-tuning. Specifically, given the trained model, we perform fine tuning. We feed a batch of images to the network, forward computing to obtain the land use prediction scores. We would like to pick the samples with distinct prediction scores to perform back propagation, and discard those samples with uniform distributed prediction scores. The intuition is that, if the prediction scores are distinct, this sample may be easier for our model to classify. On the other hand, if the prediction scores are nearly uniform distribution, it indicates that this sample could easily confuse our model. 

Let $y_{i} = [y_{i1}, y_{i2}, ..., y_{in}]$ represent the prediction (softmax) scores of training instance $i$, $n$ denotes the number of classes, which is $45$ in our situation. We calculate a probability score to determine whether to keep or discard this sample,

\begin{equation}
p_{i} = \max(0, 2 - \exp |\max(y_{i}) - \bar{y_{i}} |)
\label{eq:adaptive_prob}
\end{equation}

Here, $\bar{y_{i}}$ is the mean of prediction scores $y_{i}$. When the difference between the maximum and average of the prediction scores of one training instance is large, $p_{i}$ is low towards $0$. The loss of this training instance will be back propagated to update the model weights. Otherwise, the smaller the difference, the higher the probability of this instance being ignored. In the extreme case, when $\max(y_{i}) = \bar{y_{i}}$, $p_{i}$ will be 1, which means we would not calculate the gradients with respect to this sample. In our experiments, we set the threshold to be $0.5$ for $p_{i}$. Alternatively, we could perform soft weighting, which means we use the probability $p_{i}$ to weight the importance of each sample instead of setting a threshold to discard it. We explore both soft and hard selection scheme, but do not observe much difference. 

The positive sampling strategy is similar to the idea of hard negative mining \cite{sentiment_You_AAAI15,shrivastavaCVPR16ohem}. Hard negative mining is an useful strategy to optimize machine learning models without leveraging extra data. At the same time, it may speed up the convergence because it put more attention on the hard examples during training. But in our situation, we are not working on hard examples, but discard them. Because our dataset already contains relative amount of noisy training instances. 

In all, our training procedure has two stages. First stage, we perform a conventional end-to-end learning for land use classification. Second stage, we fine tune the trained model using the proposed online adaptive training strategy. We only back propagate the gradients with distinct prediction scores. In this way, our model learns good visual representations through the easy examples without being impacted by the noisy samples. 

\begin{table}
	\begin{center}
		\caption{Land use classification performance: both image-level classification and shapefile-level mapping accuracy. \label{tab:results_tab1}}
		\vspace{-2ex}
		\resizebox{1.0\columnwidth}{!}{%
			\begin{tabular}{ c | c || c  c  c}
				\hline
				& Classification      &    \multicolumn{3}{|c}{Mapping}    \\
				\hline
				Method			         &    Accuracy    &    Precision      &    Recall         &    F1 Score  \\
				\hline		
				\hline
				SIFT    					           &   $29.16$  &   $4.56$ 	    &    $12.85$   	&    $3.37$\\
				SIFT + Fisher Vector Encoding          &   $31.20$  &   $5.01$ 	    &    $13.67$   	&    $3.67$\\
				ResNet101 fc Layer (Pre-trained)       &   $37.87$  &   $7.92$ 	    &    $18.98$   	&    $5.59$\\
				\hline		
				\hline
				ResNet101 (Fine-Tuned)   	           &   $43.90$  &   $10.57$ 	    &    $21.67$   	&    $7.10$\\
				ResNet101 (Adaptive, Object)   &   $46.73$ 	&   $12.30$ 	    &    $25.41$   	&    $8.29$\\
				ResNet101 (Adaptive, Scene)   &   $42.93$ 	&   $10.11$ 	    &    $20.09$   	&    $6.89$\\
				ResNet101 (Two-Stream)   	           &   $\mathbf{49.54}$  &   $\mathbf{14.21}$ 	    &    $\mathbf{29.06}$   	&    $\mathbf{9.54}$\\
				\hline
			\end{tabular}
		} 
		\vspace{-2ex}
	\end{center}
\end{table} 

\subsection{Two stream: Object and Scene}
\label{subsec:two_stream}

The proposed online adaptive training can reduce the impact of noisy instances and balance the gap between Google and Flickr images during model learning. As shown in Table \ref{tab:results_tab1}, this strategy dramatically improve the performance of land use classification. However, our problem is still very challenging due to scene variations, different viewpoints and low image quality etc.

As we know, classifying land usage from user-generated images is a very high-level task. One need to understand which objects are showing, is there human in the picture, is there human-object interaction, what is the context, etc. Based on those semantic clues, we can infer what functionality this place may be used for. Inspired by multi-modal learning \cite{twostream2014,POISAR_Hu_JURSE17}, we propose a two-stream architecture: one is object-stream, the other is scene-stream.

The object-stream is the CNN model pretrained on ImageNet dataset \cite{imagenet_CVPR09}, aiming to encode object information. Most land use/cover literature use such object-centric models because we can extract good semantic features from these pretrained models. Moreover, the features generalize well across domains. However, we argue that the scene information is also important in determining land use/cover. Thus we propose to use scene-stream, a scene-centric model, to complement the object-stream and use context information to help classify land use. The scene-centric model is pretrained on Places365 dataset \cite{zhouplaces14}. Both streams are then fine tuned on our dataset as above. Implementation details are described in Section \ref{sec:implementation}. The reason we split the task to two sub-tasks is two-fold. First, the availability of two large-scale datasets lead to better pretrained CNN models. Second, we can explicitly encode object and scene information for effective land use classification due to their complementarity.

Note that, there is a possibility that the object model and scene model will collapse into one model after fine-tuning. In order to keep the complementarity between the two pre-trained models, we freeze part of the model parameters. For example, we fix the weights of the first four convolution groups of ResNet101 during fine-tuning. In this case, the low-level and mid-level features remain object- and scene- oriented. 

Our overall framework for fine-grained land use classification can be seen in Figure \ref{fig:framework_overview}.

\section{Experiments}
\label{sec:experiments}

In this section, we first describe our dataset in Section \ref{sec:dataset} and the implementation details in Section \ref{sec:implementation}. Then we report the performance of our proposed approach in Section \ref{sec:results}. We also analyze the experimental results followed by visualization and discussion. 

\subsection{Dataset}
\label{sec:dataset}
Our dataset includes two set, one is the training set, the other is the mapping set with ground truth land use types. 

The training set is constructed through keyword search on both Google images and Flickr images. The final training set consists of $2,159,460$ images spread over $45$ land use categories. Each category has around $45,000$ images and is balanced in general. We split the dataset with a ratio of $0.8$ to $0.2$ for training and validation of our CNN model. The image-level classification accuracy of our CNN model is evaluated on the validation set. 

The mapping set consists of Flickr images in the region of San Francisco over year 2016. These images has geotags. We downloaded a total of $96,382$ images, however, after geo-filtering, the final mapping set has $58,418$ images. We use the places data from Google places API as ground truth to evaluate our mapping performance. The evaluation metric is, if an image falls into the region of a shapefile, and our predicted land use type corresponds to (one of) the ground truth land use types from Google places, then we regard it as a correct land use mapping. Note that there may exist mixed land use types for a single shapefile. We adopt precision, recall and F1 score to report our mapping accuracy. Precision is calculated as the correct mappings divided by our predictions, while recall is calculated as the correct mappings divided by ground truth Google records.

\begin{figure}[t]
	\centering
	\includegraphics[width=0.9\linewidth,trim=70 200 70 205,clip]{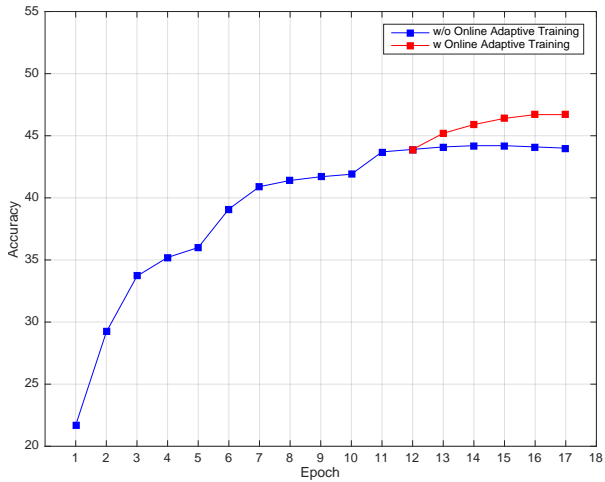}
	\caption{Image-level land use classification accuracy versus training epochs. Without online adaptive training, the accuracy come to a plateau after epoch 12 (blue curve). With the strategy, the accuracy improves further (red curve). }
	\label{fig:online_adaptive_landuse}
\end{figure}

\begin{figure*}[t]
	\centering
	\includegraphics[width=1.0\linewidth,trim=70 60 60 20,clip]{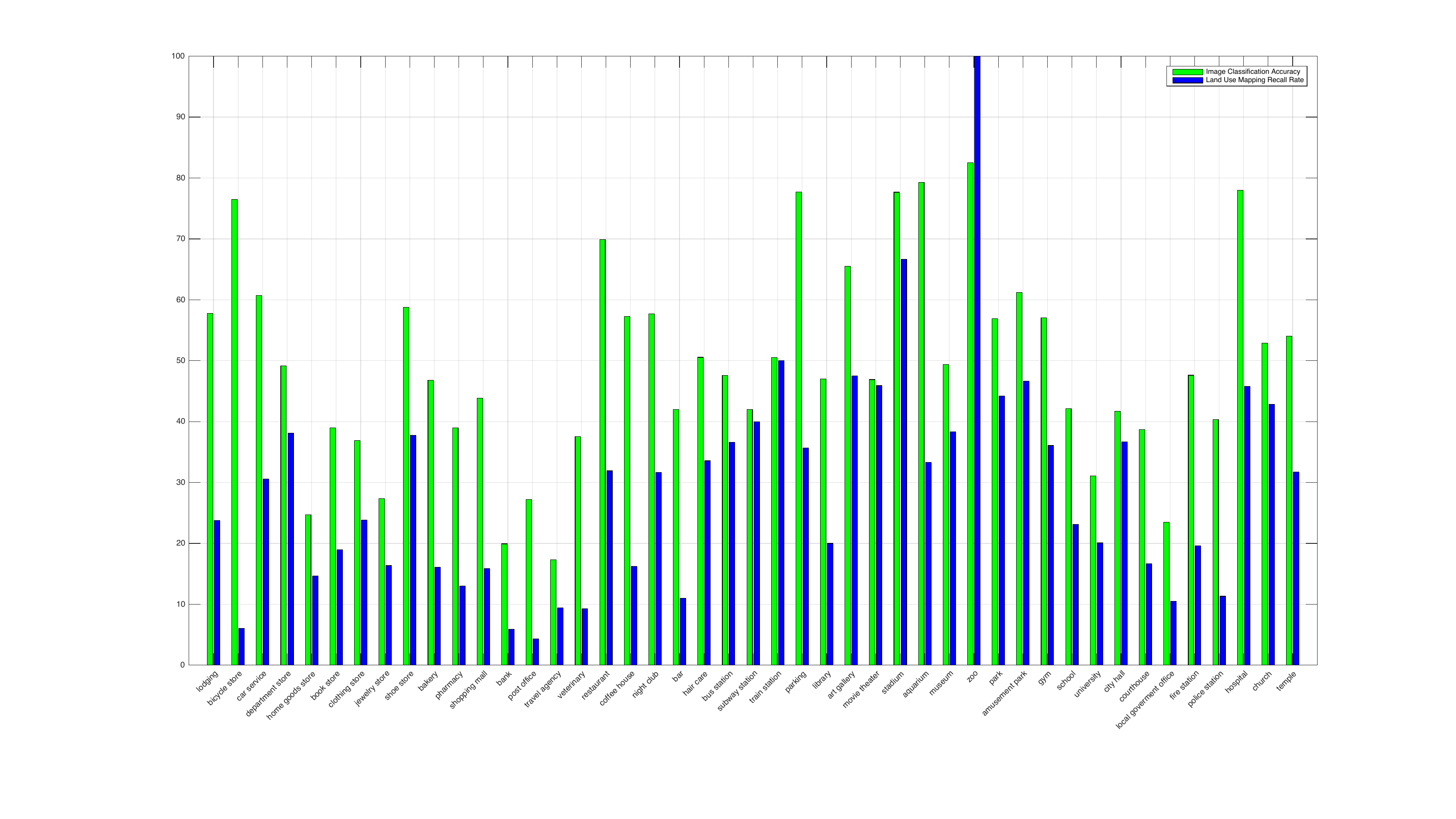}
	\caption{Per-class image-level classification accuracy (green) versus shapefile-level mapping recall (blue). The figure is best viewed in color.}
	\label{fig:image_landuse_barplot}
\end{figure*}

\subsection{Implementation Details}
\label{sec:implementation}
For the CNNs, we use the PyTorch toolbox. For all the experiments and speed evaluation, we use a workstation with an Intel Core I7 (4.00GHz) and 4 NVIDIA Titan X GPUs. 

\noindent \textbf{End-to-End learning:} 
We use ResNet101 as our network architecture, and pre-train it on ImageNet challenges \cite{imagenet_CVPR09}. During fine tuning, we change the last layer to a $45$-way classifier. The model is trained using stochastic gradient decent algorithm with the default parameter values. The batch size is set to $256$. The initial learning rate is set to $0.01$ and is divided by $10$ every $5$ epoch. We end our training at epoch $12$. 

\noindent \textbf{Online Adaptive Training} 
Given the fine-tuned model, we perform online adaptive training. We feed a batch of images to the network, forward computing to obtain the land use prediction scores. We write a custom loss layer to compute the cross entropy loss according to (\ref{eq:adaptive_prob}). We only pick the samples with distinct prediction scores to perform back propagation, and discard those samples with uniform distributed prediction scores. Since the model is already fine-tuned, the initial learning rate is set to $10^{-5}$, and divided by $10$ every one epoch. We stop the training at epoch 4. 

\noindent \textbf{Two-Stream}
We also use ResNet101 as our scene-centric model architecture, and pre-train it on Places challenges \cite{zhouplaces14}. The fine tuning details are the same as our object-centric model training. Note that, we fix part of the model parameters during fine-tuning in order to keep their complementarity. 
We perform late fusion to combine the results of our object stream and scene stream. Late fusion is weighted averaging of the predicted softmax scores from the two CNN models, the weights are set to be equal for the two models.

\subsection{Results}
\label{sec:results}
In this section, we evaluate our proposed method for fine-grained land use classification. We first compare the results from our initial end-to-end learned model with other non deep learning approaches. We then show the effectiveness of online adaptive learning strategy in the situation of using noisy web data. Finally, we indicate the advantage of combining object and scene information from two-stream networks. We also follow the experimental results with a discussion. 

The results are shown in Table \ref{tab:results_tab1}, which are evaluated on two metrics. One is the image-level classification accuracy, the other is the shapefile-level land use mapping accuracy. The image-level classification accuracy is computed based on the keyword labels, while the shapefile-level land use mapping accuracy is calculated based on the proposed ground truth land use map of San Francisco. 

\vspace{2ex}
\noindent \textbf{Image-level Classification Result:}
\vspace{1ex}

\noindent \textit{Top section of Table \ref{tab:results_tab1}}: Here we list the performance of traditional two-stage approaches. We choose SIFT as the baseline image descriptor. In addition, we compare SIFT with deep features, which are extracted from the last fully-connected layer from ResNet101 model pre-trained on ImageNet challenges. Note that the ResNet101 model here is not fine-tuned on our data, it is only used as a generic feature extractor. We can see that deep features achieves much higher image classification accuracy than traditional SIFT features. This observation is as expected because deep features are more semantic and generalized as demonstrated in other fields. We also use Fisher Vector encoding of SIFT to obtain better global features but observe limited improvement. Once we obtain the features, we use SVM as our classifier. We choose C equal to $1$ and keep other parameters as default. 

\noindent \textit{Bottom section of Table \ref{tab:results_tab1}}: Here we list the performance of our proposed method and show the improvement brought by each strategy. First, we fine tune the deep networks on our data. Despite noisy, our end-to-end trained model outperforms the above approach using the pre-trained model as a generic feature extractor. This indicates that end-to-end learning is better than two-stage method, especially during domain transfer learning. Second, we show the effectiveness of our proposed online adaptive training strategy. It improves $3\%$ over the end-to-end trained model by discarding hard examples during fine-tuning. As we can see in Figure \ref{fig:online_adaptive_landuse}, the image-level classification accuracy come to a plateau after epoch 12 (blue curve) without online adaptive training. With our proposed method, the accuracy improves further (red curve). Finally, by combining results from another stream, scene-centric model, we achieve an accuracy of $49.54\%$ on an image classification task with $45$ classes. This result is promising given how much noise does our crowd-sourcing dataset have (as shown in Figure \ref{fig:sample_images}). 

\vspace{2ex}
\noindent \textbf{Shapefile-level Mapping Result:}
\vspace{1ex}

\noindent Different from image-level classification, here, we demonstrate how image classifier trained on web images can be used to classify land use types on real map. We observe that the land use mapping performance is closely related to the image classification accuracy. The higher the image classification accuracy, the higher the precision and recall of land use mapping. Due to the large number of images we collected and their uneven spatial distribution and noisiness, the precision is low as expected because there are too many false positives. Hence we only report the recall rate in our following experiments. With good practice mentioned above, our final two-stream networks outperforms the baseline SIFT feature by $17\%$ on the challenging $45$-classes land use mapping problem.

We also show sample geo-visualizations in Figure \ref{fig:geo_visualizations}. The four regions are randomly picked from downtown San Francisco. Though each shapefile (building) may contain multiple land use types, we show the one with majority votes for clear visualization. We can see that our predictions are reasonable in most cases. For example, in the top left image, there exist many restaurants and a clothing store, which corresponds well to Google map. For the bottom right image, most of our results are correct. There is a Macy store, a hotel, and several parking structures. However, the library prediction is wrong because that building is an university. But after a sanity check, we actually have university prediction for that shapefile, it is just library has more photos and win the majority votes. Hence, we demonstrate the effectiveness of our approach both quantitatively and qualitatively. 

Note that in deep learning era, with powerful CNNs, researchers can saturate the performance on many datasets over a bunch of tasks in a fast-ever pace. However, we only achieve $29.06\%$ recall rate on our ground truth map using state-of-the-art deep network. Here emerges the two challenges, one is the inaccurate crowd-sourced data, the other is the fine granularity requirement. Similar challenges also happen in other areas, such as action recognition. \cite{sigurdsson2016hollywood} proposed a new dataset named Charades, which is a crowd-sourced fine-grained action recognition/localization benchmark. Its scale is similar to previous dataset like ActivityNet \cite{activityNet}, but state-of-the-art action recognition approach \cite{depth2action,hidden_zhu_17} only has an recognition accuracy about $15\%$, while same approach can achieve more than $70\%$ on ActivityNet. Thus, fine-grained land use mapping is a much more challenging problem than traditional land use problem in literature. We provide a ground truth map and a strong baseline to enable benchmarking future algorithms. 

\begin{figure*}[t]
	\centering
	\includegraphics[width=1.0\linewidth,trim=70 60 60 20,clip]{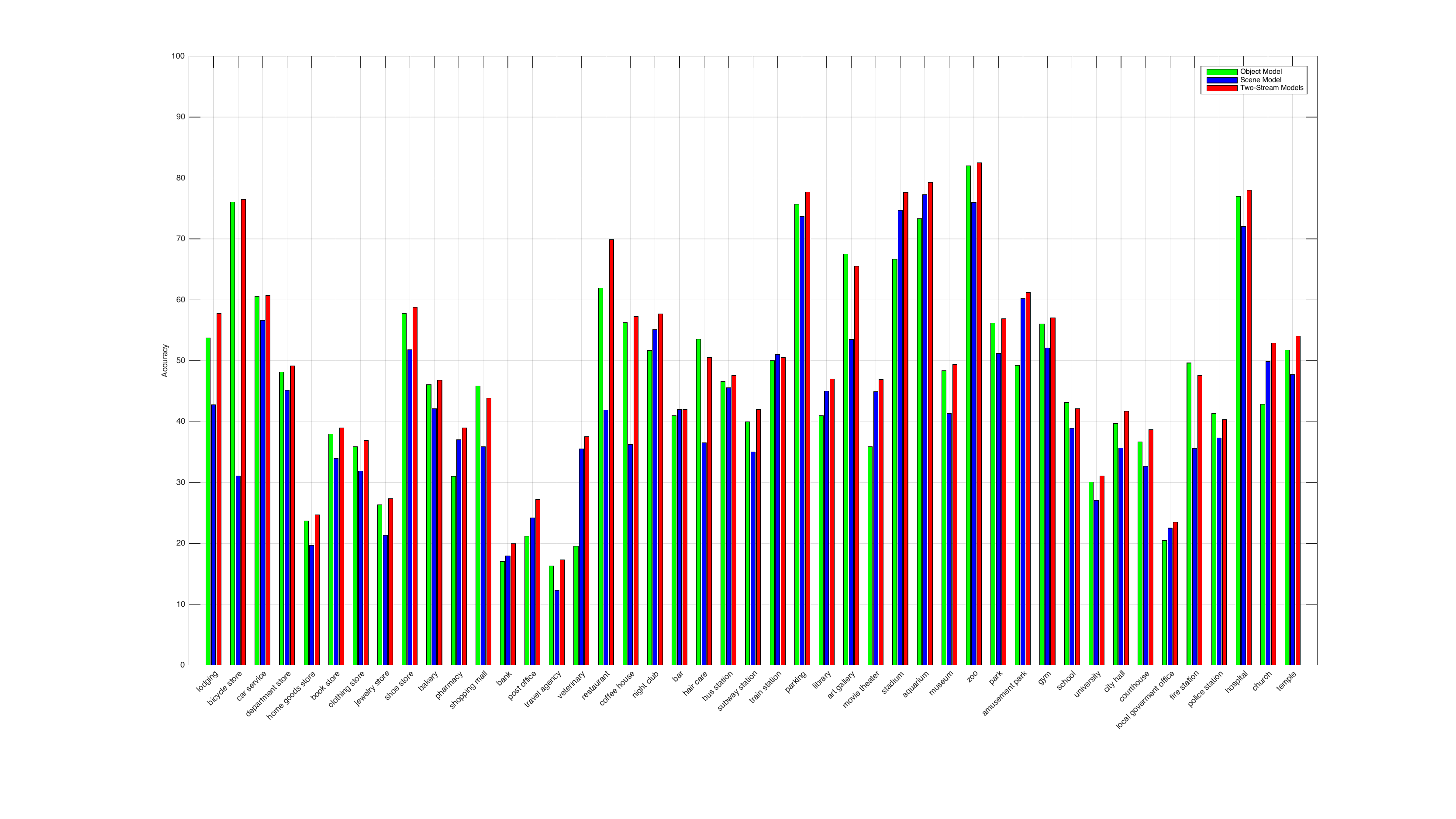}
	\caption{Per-class image-level classification accuracy of the object stream (green), scene stream (blue) and two-stream (red). We observe that object and scene information are complementary for recognizing most land use types. The figure is best viewed in color. }
	\label{fig:object_scene_barplot}
\end{figure*}

\section{Discussion}
\label{sec:discussion}

In this section, we first perform a CNN architecture search in terms of accuracy and efficiency for land use classification in Section \ref{sec:cnn_architecture_search}. Section \ref{sec:per_class_analysis} compares the per-class image-level classification accuracy with mapping recall rate, while Section \ref{sec:object_and_scene} investigates why and how object and scene model are complement. In Section \ref{sec:fine_to_coarse} we illustrate the difference between training on different dataset granularity. Finally, we use other data sources like Instagram images to demonstrate the domain adaptation ability of our method in Section \ref{sec:comparison_to_other_data_source}.

\begin{table}
	\begin{center}
		\caption{CNN architecture search. \label{tab:architecture_search}}
		\resizebox{0.7\columnwidth}{!}{%
			\begin{tabular}{  c | c | c }
				\hline
				Method									&    Accuracy (\%)  &    Speed (fps) \\
				\hline		
				\hline
				AlexNet 							&   $36.45$   &    $68.9$\\	
				VGG16 							    &   $43.88$   &    $25.6$\\	
				GoogleNet				            &   $42.03$   &    $7.8$ \\
				ResNet34 				    	    &   $43.12$   &    $19.4$ \\
				ResNet101 				    	    &   $46.73$   &    $6.4$ \\
				DenseNet121 				    	&   $47.29$   &    $3.8$ \\
				\hline
			\end{tabular}
		} 
		\vspace{-2ex}
	\end{center}
\end{table} 
	
\subsection{CNN Architecture Search}
\label{sec:cnn_architecture_search}

As we know, CNN architecture is crucial for its performance on different tasks, e.g., the depth, width and the internal connections of the CNN. We perform a CNN architecture search to find the best network for classifying land use types given noisy web images in terms of the trade-off between accuracy and efficiency. Here, we compare several architectures, AlexNet, VGG16, GoogleNet, ResNet34, ResNet101 and DenseNet121. These architectures are well designed and have been widely used in many areas. 

The results can be seen in Table \ref{tab:architecture_search}. Here, we only report the image classification accuracy of the object-centric model. 
The speed is evaluated using frame per second (fps) metric. The higher the fps, the faster the model runs. 

In general, we obtain better performance with deeper networks. There is one interesting observation that VGG16 performs better than ResNet34 despite that VGG16 has $16$ layers while ResNet34 has $34$ layers. In addition, ResNet34 performs better than VGG16 on object recognition task in ImageNet challenges \cite{imagenet_CVPR09}. This demonstrates that VGG16 is a more robust model which has good generalization towards noisy dataset. 

DenseNet121 performs the best due to its deeper network, implicit supervision and being less prone to overfitting. However, it is both memory and time consuming. Considering the trade-off between accuracy and efficiency, we choose ResNet101 as our CNN architecture. 

\subsection{Image Classification versus Land Use Mapping}
\label{sec:per_class_analysis}
In Section \ref{sec:results}, we find that land use mapping performance is closely related to the image classification accuracy. Intuitively, this makes sense because a better image classification model will lead to better performance for the following task. Here, we explore their relationships in details by comparing the per-class accuracy. The results can be seen in Figure \ref{fig:image_landuse_barplot}.

We observe that the recall rate and image classification accuracy for some categories do not correlate. For example, the ``bicycle store'' class, the image classification accuracy is almost $80\%$ because our object model can easily detect bicycles to determine its land use. However, when we do the mapping, the recall rate is lower than $10\%$. The reason maybe that the bicycle stores are usually small places, the noisiness of the web images and the inaccurate geotag will lead to bad mapping performance. For some other land use types, like ``bank'', ``local government office'', ``courthouse'' and ``library'', the number of photos used for mapping is quite small due to privacy issue or photographer intent. Thus their land use recall rate is also very low. 

In general, a better land use image classification model will lead to better land use mapping performance, but this may not be the case for some land use types. Fine-grained land use classification remains challenging due to a variety of reasons, such as uneven spatial distribution, inaccurate geotags, low image quality, lack of images for some land use types etc. It is a problem beyond pure pixel-level image understanding, but requires human common knowledge.  

\subsection{Object and Scene}
\label{sec:object_and_scene}
As evidenced in Table \ref{tab:results_tab1}, two-stream networks outperform each single of them, which indicates that object and scene stream should be complementary to each other. In this section, we further investigate why and how they are complement.  

We compare the per-class accuracy of the object model, scene model and two-stream models as in Figure \ref{fig:object_scene_barplot}. We make two observations: (i) Object model obtains better performance for most land use types than scene model. This maybe because land use is determined by how people use the land, and objects are usually more indicative than scenes. For example, object model outperforms scene model on class ``bicycle stores'' by a large margin because the existence of a bicycle is important to determine its land use. However, the surrounding scene of a bicycle store may be similar to other land use types. (ii) Object and scene model are complementary to each other. $39$ out of $45$ classes obtain better performance due to late fusion of the two models. For the other $6$ classes, their performance decrease are quite marginal. 

The top $5$ classes with the most improvement (amount in parenthesis) after incorporating scene model are ``veterinary'' ($18.12\%$), ``amusement park'' ($12.30\%$), ``movie theater'' ($11.04\%$), ``stadium'' ($10.97\%$) and ``church'' ($9.86\%$). We believe that the scene cue is important for the recognition of these land use types because there are no specific objects that are related to the classes.  

The $6$ classes that has decreased performance are ``hair care'' ($-3.77\%$), ``fire station'' ($-2.49\%$), ``shopping mall'' ($-2.31\%$), ``art gallery'' ($-2.04\%$), ``school'' ($-1.68\%$) and ``police station'' ($-1.20\%$). The reason is because the performance of our scene model on these 6 classes are not promising which lower the overall performance. 

\begin{table}
	\begin{center}
		\caption{Image-level classification accuracy on different dataset granularity. \label{tab:fine_to_coarse}}
		\resizebox{0.8\columnwidth}{!}{%
			\begin{tabular}{  c | c | c | c}
				\hline
				Method									&    45-way   &    16-way  &    5-way \\
				\hline		
				\hline
				Fine Granularity 							&   $\mathbf{46.7}$   &    $\mathbf{61.8}$  &    $\mathbf{75.6}$\\	
				Middle Granularity  &   $-$   &    $60.2$ &    $68.4$\\		
				Coarse Granularity  &   $-$   &    $-$ &    $49.3$\\
				\hline
			\end{tabular}
		} 
		\vspace{-2ex}
	\end{center}
\end{table}

\subsection{Fine to Coarse}
\label{sec:fine_to_coarse}
In some circumstances, we do not need land use maps with such fine granularity. The $16$ middle classes or the $5$ top classes will meet our requirements. There is a question that should we train a model with fine granularity and combine the results later to get higher level predictions, or we directly train a model with coarse granularity. Here, we investigate this issue by training another 16-way and 5-way classifier on our dataset. 

We call our $45$-way classifier as fine granularity model, $16$-way classifier as middle granularity model and $5$-way classifier as coarse granularity model. The results are shown in Table \ref{tab:fine_to_coarse}. We can see that training on fine granularity is beneficial. The model can learn discriminant features for different land use types. We can achieve an accuracy of $61.8\%$ for $16$ middle classes and $75.6\%$ for $5$ top classes. The middle granularity model performs worse than fine granularity model because these $16$ classes may correlate in pixel space. A single model can not differentiate them without human common knowledge. The coarse granularity model performs the worst, with an accuracy of only $49.3\%$, slightly better than random guess. This is because the images within one class are so different that will easily confuse the model. For example, the ``General sales or services'' top class includes many concepts that are visually quite different, like bank and bakery, hair care and restaurant. The model may learn nothing given such serious intra-class variance. 

\subsection{Comparison to Other Data Source}
\label{sec:comparison_to_other_data_source}

In this section, we aim to explore using other data source like Instagram images to evaluate the land use mapping performance. On one hand, this will show the robustness of our proposed method and good transfer learning capability. On the other hand, we show that the ground truth land use map of San Francisco can serve as a benchmark to evaluate other algorithms using various data sources. 

We download a total of $121,567$ images within the city of San Francisco in the year of $2014$ using Instagram API\footnote{We pick year 2014 instead of 2016 is because Instagram recently limits its API from massive downloading of user images with location information.}. We directly use the trained models to classify each image to its predicted land use types. The recall rate of the land use mapping performance is $17.3\%$, although lower than the best result $29.03\%$, we demonstrate good domain adaptation ability of our model. The accuracy is similar to $18.9\%$, that of using pre-trained deep CNN models as generic feature extractor. Note that, Instagram images are quite different compare to Flickr images in style. Most Instagram images are selfie or selfie-like, which describes near-sight scenes. This is usually not useful towards recognition of land use types. 
We could use other data sources, like tweets or Youtube videos, to further evaluate our method, but we leave it for future work. 

\section{Conclusion}
\label{sec:conclusion}

We introduce a ground truth fine-grained land use map in city-scale and present a framework for land use classification using georeferenced ground-level images. Our dataset structure has a 3-level hierarchy, with 5 top classes, 16 middle classes and 45 bottom classes. To the best of our knowledge, the number of land use types in finest granularity, $45$, is much more diverse than previous literature. Our two-stream models, together with the online adaptive training strategy, achieve promising results on the challenging 45-way land use classification problem. We believe our provided ground truth map can encourage further research on fine-grained land use classification and our results serve as a strong baseline.

In the future, we would like to improve our two-stream models in the following directions. First, we plan to explore multi-model information, such as text, audio, video or various input signals, to investigate their complementarity. Second, we adopt the idea of human-in-the-loop since fine-grained land use mapping problem is a very challenging problem. We probably need human knowledge to produce accurate land use map for governmental or industrial use such as city zoning map. Third, based on the observation that object-stream usually achieve better performance than scene-stream, we could explore further on object stream by applying off-the-shelf object detectors. Such local approach may achieve better performance and provide evidence for which object(s) are the key factor to determine the land use. 


\bibliographystyle{IEEEtran}
\bibliography{IEEEabrv,journalBIB}


\appendices
\section{Full Dataset Hierarchy}

We list the full dataset hierarchy in 3-level as below. There are $5$ top classes, $16$ middle classes and $45$ bottom classes. 

\begin{enumerate}
    \item Residence or accommodation functions
    \begin{enumerate}
        \item Hotels, motels, or other accommodation services
        \begin{itemize}
            \item lodging
        \end{itemize}
    \end{enumerate}
    \item General sales or services
    \begin{enumerate}
        \item Retail sales or service
        \begin{itemize}
            \item bicycle$\_$store
            \item car$\_$service
            \item department$\_$store
            \item home$\_$goods$\_$store
            \item book$\_$store
            \item clothing$\_$store
            \item jewelry$\_$store
            \item shoe$\_$store
            \item bakery
            \item pharmacy
            \item shopping$\_$mall
        \end{itemize}
        \item Finance and Insurance
        \begin{itemize}
            \item bank
        \end{itemize}
        \item Business, professional, scientific, and technical services
        \begin{itemize}
            \item post$\_$office
            \item travel$\_$agency
            \item veterinary$\_$care
        \end{itemize}
        \item Food services
        \begin{itemize}
            \item restaurant
            \item coffee$\_$house
            \item night$\_$club
            \item bar
        \end{itemize}
        \item Personal services
        \begin{itemize}
            \item hair$\_$care
        \end{itemize}
    \end{enumerate}
    \item Transportation, communication, information, and utilities
    \begin{enumerate}
        \item Transportation service
        \begin{itemize}
            \item bus$\_$station
            \item subway$\_$station
            \item train$\_$station
            \item parking
        \end{itemize}
        \item Communications and information
        \begin{itemize}
            \item library
        \end{itemize}
    \end{enumerate}
    \item Arts, entertainment and recreation
    \begin{enumerate}
        \item Performing arts or supporting establishment
        \begin{itemize}
            \item art$\_$gallery
            \item movie$\_$theater
            \item stadium
        \end{itemize}
        \item Museums and other special purpose recreational institutions
        \begin{itemize}
            \item aquarium
            \item museum
            \item zoo
        \end{itemize}
        \item Amusement, sports, or recreation establishment
        \begin{itemize}
            \item park
            \item amusement$\_$park
            \item gym
        \end{itemize}
    \end{enumerate}
    \item Education, public admin, health care and other institution
    \begin{enumerate}
        \item Educational services
        \begin{itemize}
            \item school
            \item university
        \end{itemize}
        \item Public administration
        \begin{itemize}
            \item city$\_$hall
            \item courthouse
            \item local$\_$government$\_$office
        \end{itemize}
        \item Public safety
        \begin{itemize}
            \item fire$\_$station
            \item police$\_$station
        \end{itemize}
        \item Health and human services
        \begin{itemize}
            \item hospital
        \end{itemize}
        \item Religious institutions
        \begin{itemize}
            \item church
            \item temple
        \end{itemize}
    \end{enumerate}
\end{enumerate}

\end{document}